\documentclass[sigconf]{acmart}
\usepackage{newfloat}
\usepackage{mathtools}
\usepackage{listings}
\usepackage{multirow}
\usepackage{float}
\usepackage{caption}
\usepackage{subfigure}
\usepackage{graphicx}
\usepackage{stfloats}
\usepackage{color}
\usepackage{fontawesome}

\usepackage{bbding}
\AtBeginDocument{%
  \providecommand\BibTeX{{%
    \normalfont B\kern-0.5em{\scshape i\kern-0.25em b}\kern-0.8em\TeX}}}

\copyrightyear{2022}
\acmYear{2022}
\setcopyright{acmcopyright}
\acmConference[SIGIR '22] {Proceedings of the 45th International ACM SIGIR Conference on Research and Development in Information Retrieval}{July 11--15, 2022}{Madrid, Spain}
\acmBooktitle{Proceedings of the 45th International ACM SIGIR Conference on Research and Development in Information Retrieval (SIGIR '22), July 11--15, 2022, Madrid, Spain}
\acmPrice{15.00}
\acmISBN{978-1-4503-8732-3/22/07}
\acmDOI{10.1145/3477495.3531930}

\begin{document}
\fancyhead{}

\title{A Weakly Supervised Propagation Model for Rumor Verification and Stance Detection with Multiple Instance Learning}


\author{Ruichao Yang}
\affiliation{%
  \institution{Hong Kong Baptist University}
  \city{Hong Kong SAR}
  \country{China}
}
\email{csrcyang@comp.hkbu.edu.hk}

\author{Jing Ma\textsuperscript{\Envelope}}
\affiliation{%
  \institution{Hong Kong Baptist University}
  \city{Hong Kong SAR}
  \country{China}
}
\email{majing@comp.hkbu.edu.hk}

\author{Hongzhan Lin}
\affiliation{%
  \institution{Beijing University of Posts and Telecommunications}
  \city{Beijing}
  \country{China}
}
\email{linhongzhan@bupt.edu.cn}

\author{Wei Gao}
\affiliation{%
   \institution{Singapore Management University}
   \country{Singapore}
}
\email{weigao@smu.edu.sg}






\renewcommand{\shortauthors}{Yang, et al.}

\begin{abstract}
  The diffusion of rumors on social media generally follows a propagation tree structure, which provides valuable clues on how an original message is transmitted and responded by users over time. Recent studies reveal that rumor verification and stance detection are two relevant tasks that can jointly enhance each other despite their differences. For example, rumors can be debunked by cross-checking the stances conveyed by their relevant posts, and stances are also conditioned on the nature of the rumor. However, stance detection typically requires a large training set of labeled stances at post level, which are rare and costly to annotate.
  

Enlightened by Multiple Instance Learning (MIL) scheme, we propose a novel weakly supervised joint learning framework for rumor verification and stance detection which only requires bag-level class labels concerning the rumor's veracity. Specifically, based on the propagation trees of source posts, we convert the two multi-class problems into multiple MIL-based binary classification problems where each binary model is focused on differentiating a target class (of rumor or stance) from the remaining classes. Then, we propose a hierarchical attention mechanism to aggregate the binary predictions, including (1) a bottom-up/top-down tree attention layer to aggregate binary stances into binary veracity; and (2) a discriminative attention layer to aggregate the binary class into finer-grained classes.
Extensive experiments conducted on three Twitter-based datasets demonstrate promising performance of our model on both claim-level rumor detection and post-level stance classification compared with state-of-the-art methods.
\end{abstract}



\maketitle

\section{Introduction}
The rapid development of social networks has spawned a large number of rumors, which jeopardize the environment of online community and result in harmful consequences to the individuals and our society. For instance, during the COVID-19 pandemic, a false rumor claimed that ``magnetism will be generated in the body after the injection of coronavirus vaccine"\footnote{\url{https://www.bbc.com/news/av/57207134}} went viral and shared millions of times on social media platforms, which causes vaccination hesitation to the public and delays the establishment of the herd immunity in the population. 
It is meanwhile noteworthy that a wide variety of online opinions spread about a rumor  
can be useful for us to distill the collective wisdom of crowd in recognizing some very challenging rumors~\cite{zubiaga2018detection}. 
In recent years, this has inspired researchers to develop automatic rumor verification approaches by leveraging large-scale analysis of online posts to mitigate the harm of rumors.
\begin{figure*}[htbp]
\centering
\includegraphics[width=5.6in]{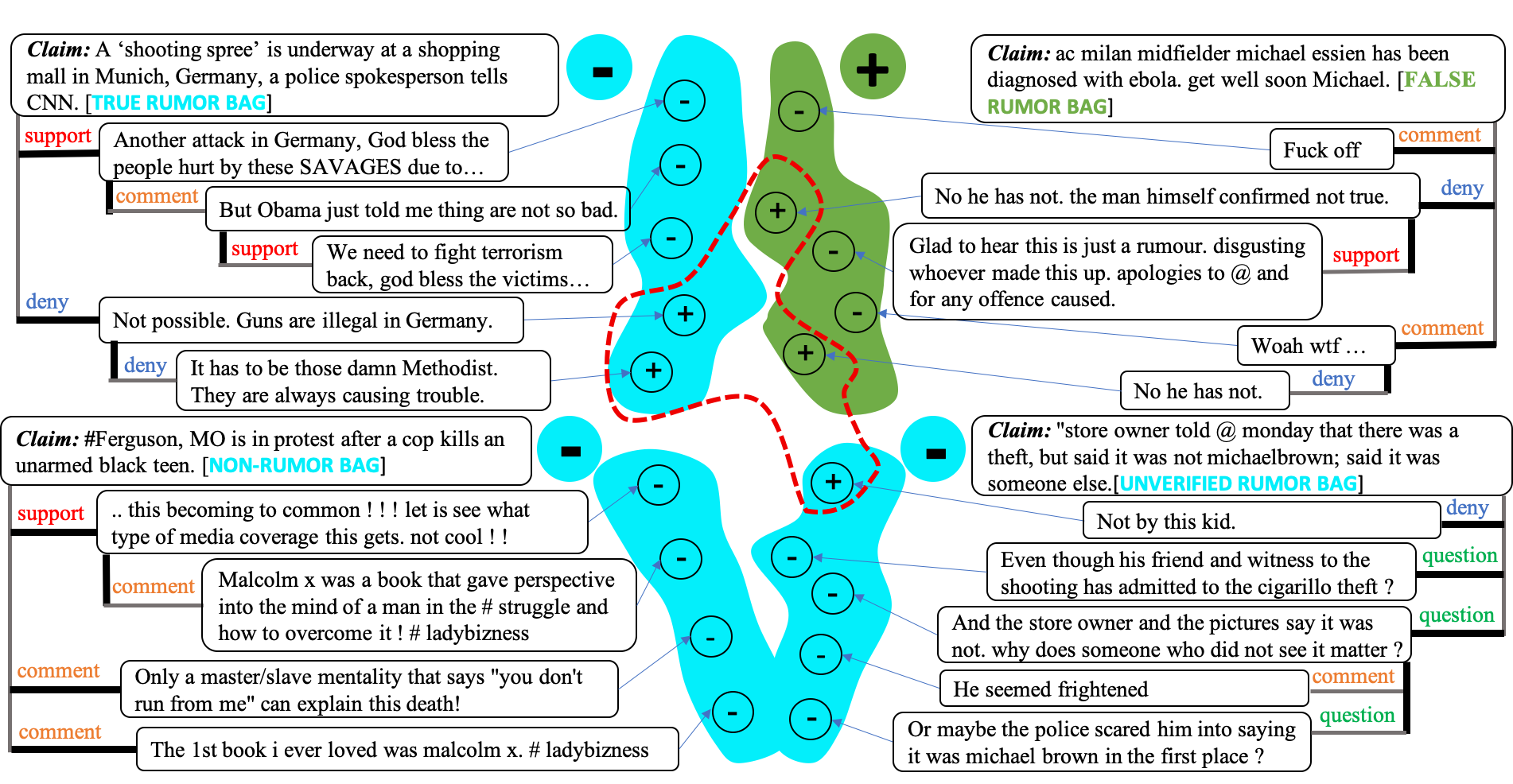}
\caption{An illustration of tree-based MIL binary classification for simultaneous rumor and stance detection.}
\label{fig:fig1}
\end{figure*}

Rumor verification is a task to determine the veracity of a given claim about some subject matter~\cite{li2019rumor}. 
Most rumor verification methods are focused on training a supervised model by utilizing pre-defined features~\cite{castillo2011information,liu2015real,yang2012automatic} or enquiring patterns~\cite{zhao2015enquiring} over the claim and its responding posts.
To avoid the painstaking feature engineering, neural methods such as Recurrent Neural Networks (RNNs)~\cite{ma2016detecting} and Convolutional Neural Networks (CNNs)~\cite{yu2017convolutional} were proposed to learn discriminative features from the sequential structure of rumor propagation. 
To further capture the complex propagation patterns, kernel learning algorithms were designed to compare different propagation trees~\cite{wu2015false,ma2017detect,rosenfeld2020kernel}. Propagation trees were also utilized to guide feature learning for classifying different types of rumors on Twitter based on Recursive Neural Networks (RvNN)~\cite{ma2018rumor} or Transformer-based models~\cite{khoo2020interpretable,ma2020debunking}. 

Stance detection on social media aims to determine the attitude expressed in a post towards a specific target. Previous studies conducted massive manual analysis on stances pertaining to different rumor types~\cite{mendoza2010twitter}. Subsequent methods leveraged temporal traits to classify rumor stances with Gaussian Process~\cite{lukasik2015classifying} and Hawkes Process~\cite{lukasik2016hawkes}. Some studies proposed to train supervised models based on hand-crafted features~\cite{zeng2016unconfirmed,bhatt2018combining,zubiaga2018discourse}. 
To alleviate feature engineering, \citet{zhang2019stances} proposed to learn a hierarchical representation of stance classes to overcome the class imbalance problem. Further, multi-task learning framework was utilized to mutually reinforce stance detection and rumor classification simultaneously~\cite{ma2018detect,kochkina2018all,li-etal-2019-rumor-detection,wei2019modeling}. 
However, they generally require a large stance corpus annotated at post level for model training, which is a daunting issue. 
While some unsupervised methods were proposed~\cite{kobbe2020unsupervised,allaway2020zero}, they achieved poor generalizability due to the specifically crafted features or pre-trained task-specific models, which cannot be robustly migrated to other tasks. 


Past research revealed that there is close relationship between rumor verification and stance classification tasks. For instance, the voices of opposition and doubt seem to always appear among the posts along with the spread of rumors~\cite{mendoza2010twitter}; also, there are many skepticism and enquiries posts sparked to find out the claim's veracity~\cite{zubiaga2015towards}. Moreover, propagation structure provides valuable clues regarding how a claim is transmitted and opinionated by users over time~\cite{ma2020debunking,wei2019modeling}. Figure~\ref{fig:fig1} exemplifies the dissemination of four types of rumorous claims, where each claim is represented as a propagation tree consisting of the source post and user responses. 
We observe that: (1) a response generally reacts towards its responded post instead of the source claim (i.e., the claim) directly; (2) a post denying a false rumor claim tends to trigger supporting replies to confirm the objection; and (3) a post denying a true rumor claim can generally spark denying replies that object the denial. 
So the stances expressed in such structured responses may contain signals for distinguishing rumor classes. On the other hand, we can observe that given the rumor veracity class, false rumors tend to attract more denying posts and unfold with more edges like ``$support \to deny$" and ``$deny \to support$" than true claims, while unverified rumors contain more questioning posts with more ``$comment \to comment$" and ``$comment \to question$" edges than the other classes of claims. Intuitively, we can deduce the stances of the individual posts according to the rumor veracity and response structure by first determining the stances for some salient nodes directly, and further for the nodes in connection with the nodes with the help of these featured edges.
In this paper, we propose a weakly supervised approach, which is an extension of Multiple Instance Learning (MIL)~\cite{foulds2010review}, jointly detect stance and verify rumors only using the veracity labels of rumorous claims (i.e., so-called bag-level annotation). 
MIL-based classifier is traditionally used to classify individual instances (e.g., sentences) in a bag (e.g., document) and then deduce the bag-level prediction by aggregating the instance-level prediction while using only bag-level annotations for a weak supervision in the training. The original MIL assumes that the class labels defined at these two levels are binary and homogeneous. In this work, however, the two tasks have their own multi-class labels defined independently. The main challenge lies in how to correlate the two different sets of class labels and turn the problem to become learnable with the existing MIL framework. Another challenge is to make inference from one set of classes to another set based on the feature representations learned from propagation trees which is the essential propagation structure of social media posts. 

To this end, we firstly convert the multi-class classification problem into multiple MIL-based binary problems. For an illustration, Figure~\ref{fig:fig1} shows the idea of our joint prediction model, where we can treat False rumor and Deny stance (F-D pair) as the positive class (i.e., the target class) at the bag and instance level, respectively, while regarding the rest of the classes as negative. Provided that rumor veracity labels contain False rumor (F), True rumor (T), Unverified rumor (U) and Non-rumor (N) while post stance labels consist of Deny (D), Support (S), Question (Q) and Comment (C), quite a few possible veracity-stance class pairs can exist, and each veracity-stance pair can indicate the target class of a claim and a responsive post for the binary classification on the claim veracity and post stance, respectively. We thus develop MIL-based top-down/bottom-up propagation models for representation learning and classification, corresponding to different edge directions. 
In order to synthesize the binary results of the set of binary veracity-stance prediction models, we propose a novel hierarchical attention mechanism to (1) aggregate the obtained stances from each binary MIL model for inferring the rumor claim veracity; and (2) combine the multiple binary results into a unified result of multiple classes. 
In this way, the stance of each post is obtained by attending over the the predictions of all the binary (instance-level) stance classifiers. Similarly, we predict the rumor class of each claim by attending over the binary (bag-level) veracity predictions following the weighted collective assumption in MIL~\cite{foulds2010review}, which is a variant of standard MIL assuming that the bag-level label relies more on vital instances. 
Extensive experiments conducted on three Twitter-based benchmark datasets demonstrate that our MIL-based method achieves promising results for both rumor verification and stance detection tasks.

\section{Related Work}
In this section, we provide a brief review of the research works on three different topics that are related to our study.

\textbf{Rumor Verification.} 
Pioneer research on automatic rumor verification was focused on pre-defined features or rules crafted from texts, users, and propagation patterns to train supervised classification models~\cite{qazvinian2011rumor,liu2015real,yang2012automatic}. \citet{jin2016news} exploited the conflicting viewpoints in a credibility propagation network for verifying news stories propagated via tweets.
To avoid feature engineering, subsequent studies proposed data-driven methods based on neural networks such as RNN~\cite{ma2016detecting} and CNN~\cite{yu2017convolutional} to automatically capture rumor-indicative patterns. Considering the close correlations among rumor and stance categories, multi-task learning framework was utilized to mutually reinforce rumor verification and stance detection tasks~\cite{ma2018detect,kochkina2018all,li-etal-2019-rumor-detection,wei2019modeling}.
To model complex structures of post propagation, more advanced approaches were proposed to use tree kneral-based method~\cite{wu2015false,ma2017detect}, tree-structured RvNN~\cite{ma2018rumor,wei2019modeling}, hybrid RNN-CNN model~\cite{liu2018early}, variants of Transformer~\cite{khoo2020interpretable,ma2020debunking}, graph co-attention networks~\cite{lu2020gcan} and Bi-directional Graph Neural Networks (BiGCN)~\cite{bian2020rumor, lin2021rumor} for classify different kinds of rumors. 
Inspired by their success, we develop our approach based on the tree-structured model. 

\textbf{Stance Detection.} 
Manual analysis on stance revealed that there are close relationship between specific veracity categories and stances~\cite{mendoza2010twitter}. 
In the follow-up studies, a range of hand-crafted features~\cite{zeng2016unconfirmed,bhatt2018combining,zubiaga2018discourse} as well as temporal traits~\cite{lukasik2015classifying,lukasik2016hawkes} were studied to train stance detection models. 
More recently, deep neural networks were utilized for stance representation learning and classification to alleviate feature engineering and pursue stronger generalizability, 
such as bidirectional RNNs~\cite{augenstein2016stance} and two-layer neural networks for learning hierarchical representation of stance classes~\cite{zhang2019stances}. 
Some studies further took into account conversation structure, such as the tree-based LSTM model for detecting stances~\cite{zubiaga2016stance,kochkina2017turing} and the tree-structured multi-task framework for joint detection of rumors and stances~\cite{wei2019modeling}. 
However, these methods require a large stance corpus annotated at post level for training. While several unsupervised stance models based on pre-defined rules~\cite{kobbe2020unsupervised} and pre-trained models~\cite{allaway2020zero} were available, their generalizability on stance detection is always a critical concern due to different problem settings. In this paper, we propose a weakly supervised propagation model to classify rumor categories and post stances simultaneously with no need of post stance labels. 

\textbf{Multiple Instance Learning (MIL).} MIL is a type of weakly supervised learning which aims to learn a classifier with coarse-grained bag-level annotation to assign labels to fine-grained instances, where instances are arranged in the bag~\cite{dietterich1997solving}. 
Further research following MIL proposed some variants by extending the threshold-based, count-based and weighted collective MIL assumptions to infer the bag-level label~\cite{foulds2010review}. In recent years, MIL has been successfully applied to various applications in the field of Natural Language Processing (NLP), such as 
the unified MIL framework that simultaneously classified news articles and extracted sentences~\cite{wang2016multiple}, 
the MIL-based model for user personalized satisfaction prediction in Community Question Answering (CQA)~\cite{chen2017user}, the context-assisted MIL method using service dialogues for customer satisfaction analysis in E-Commerce~\cite{song-etal-2019-using}, and the attention-based MIL network for fashion outfit recommendation~\cite{lin2020outfitnet}, etc. 
However, the original MIL is specifically designed for binary classification for the instances without complex structures, and the bag-level labels should be compatible with the instance-level labels. In this paper, we design a tree-based MIL framework to convert the multi-class problem into multiple binary classifiers and solve the issue of incompatible labels between the bag and instance levels.

\begin{figure*}[htbp]
\centering
\includegraphics[width=6in]{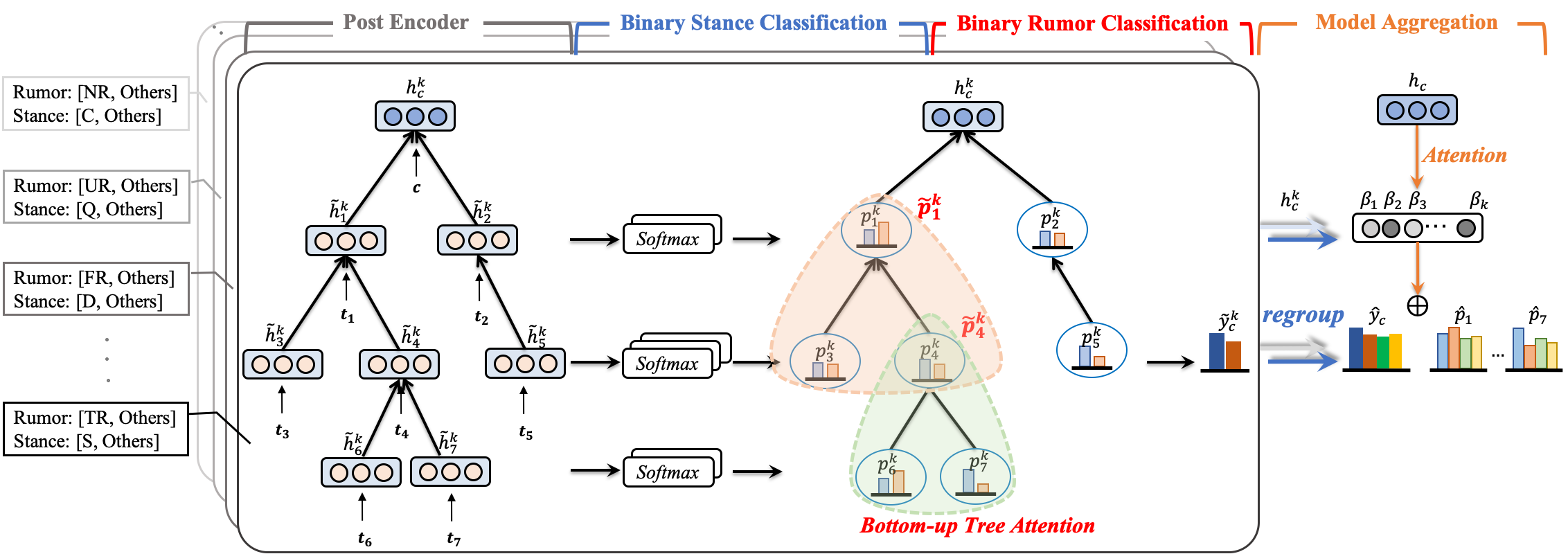}
\caption{A framework of MIL-based model with bottom-up tree. The edge direction in the tree corresponds to stance feature aggregation recursively from bottom to up. 
$\tilde{p}_i^k$ denotes the aggregated stance in a subtree rooted at $t_i$.}
\label{fig:Model_bu}
\end{figure*}
\begin{figure*}[htbp]
\centering
\includegraphics[width=6in]{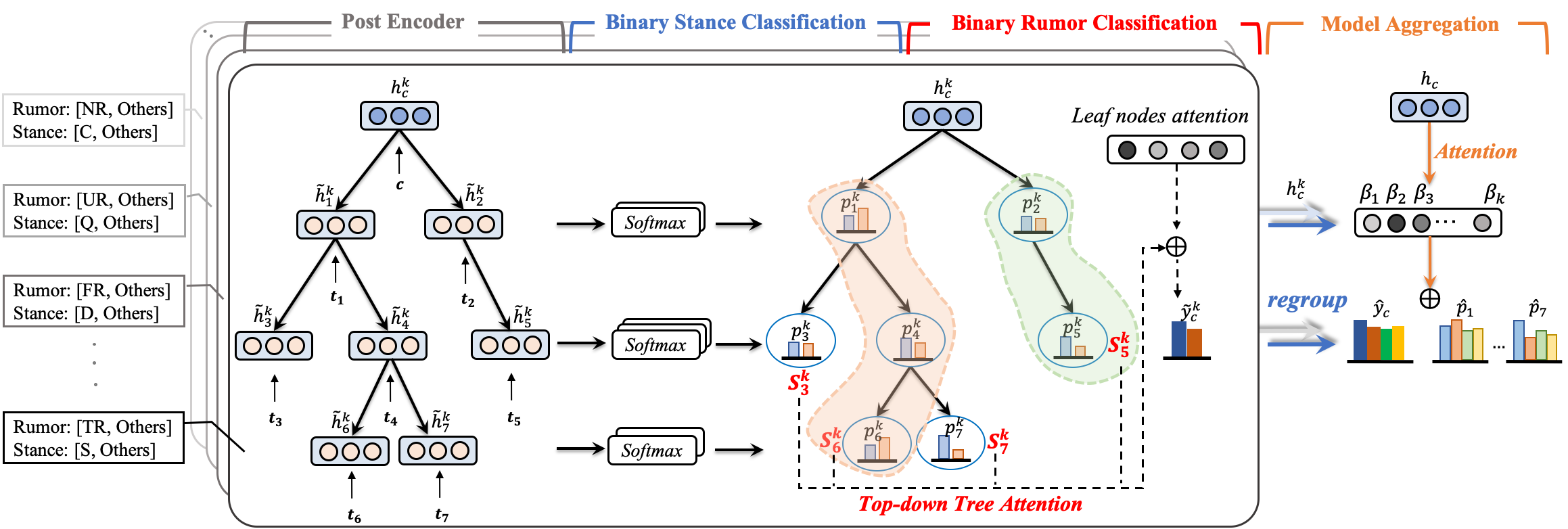}
\caption{A framework of MIL-based model with top-down tree. 
$S_l^k$ denotes the aggregated stance for specific path from $c$ to $t_l$. Attending over the updated representation of all Leaf nodes correspond to selecting more informative propagation paths.}
\label{fig:Model_td}
\end{figure*}

\section{Problem Statement} \label{sec:ps}
We define a rumor dataset as $\{\mathcal{C}\}$, where each training instance $\mathcal{C}=(c,X,y)$ is a tuple consisting of a claim $c$, a sequence of relevant tweets $X=(t_1, t_2, \cdots, t_T)$ and a veracity label $y$ of the claim. Note that although the tweets are presented in a chronological order, there are explicit connections such as response or repost relations among them. Inspired by~\citet{ma2018rumor}, we represent each claim as two different propagation trees with distinct edge directions: (1) \textit{Bottom-up tree} where the responding nodes point to their responded nodes, similar to a citation network; 
and (2) \textit{Top-down tree} where the edges follow the direction of information diffusion by reversing the edge direction in the Bottom-up tree. 
%
We define the following two tasks:
\begin{itemize}
\item \textbf{Stance Detection:} The task is to determine the post-level stance $p_i$ for a post $t_i$ expressed over the veracity of claim $c$. That is, $f: t_1 t_2 \dots t_T, c \to p_1 p_2 \cdots p_T$, where $p_i$ is the stance label that takes one of Support (S), Deny (D), Question (Q) or Comment (C). 
Here the Comment is assigned to the posts that do not have clear orientations to the claim's veracity.
\item \textbf{Rumor Verification:} The task is to classify the claim $c$ on top of the post stances as one of the four possible veracity labels $y$: Non-rumor (N), True rumor (T), False rumor (F) or Unverified rumor (U). That is, $g: p_1 p_2 \cdots p_T \to y$, where $\{p_1 p_2 \cdots p_T\}$ follows a 
top-down/bottom-up propagation tree structure. In our approach, we hypothesize that the task can be modeled following the weighted collective assumption of MIL in a sense that the bag-level label relies more on vital instances~\cite{foulds2010review}. In our task, bags (i.e., claims) are predictively labeled with the ``most likely" class according to the distribution of instances (i.e., posts) labels in the tree.
\end{itemize}

\section{Our Approach}

In this section, we will describe our extension to the original MIL framework for joint rumor verification and stance detection based on the bottom-up and top-down propagation tree structures. Due to the fine-grained categories of stance and rumor veracity, we first decompose the multi-class problem into multiple binary classification problems and then aggregate the predicted binary classes into multi-way classes. Without the loss of generality, we assume the number of rumor classes is $N_r$ and that of the stance classes is $N_s$, and thus there can be $K=N_r*N_s$ possible veracity-stance target class pairs corresponding to $K$
binary classifiers weakly supervised for detecting rumor veracity and post stance simultaneously.


\subsection{Post Encoding}
We represent each post as a word sequence $t_i=\{w_{i,1} w_{i,2} \cdots w_{i,|t_i|}\}$, where $w_{i,j} \in \mathbb{R}^d$ is a $d$-dimensional vector that can be initialized with pre-trained word embeddings. 
We map each $w_{i,j}$ 
into a fixed-size hidden vector using standard GRU~\cite{cho2014properties} and obtain the post-level vector for $c$ and $t_i$ by using two independent GRU-based encoders\footnote{Here we choose GRU-based encoder because GRU needs fewer parameters than LSTM. The GRU-based encoder can be replaced with pre-trained language models such as ELMO~\cite{peters2018deep}, BERT~\cite{devlin2019bert}, Roberta~\cite{liu2019roberta}, and BERTweet~\cite{nguyen2020bertweet}.
}: 
\begin{equation}\label{equ:GRUPostRepresentation}
\begin{split}
    & h_{c} = h_{|c|}=GRU(w_{|c|}, h_{|c|-1}, \theta_{c}) \\
    & h_{i} = h_{|t_i|}=GRU(w_{|t_i|}, h_{|t_i|-1}, \theta_{X})
\end{split}
\end{equation}
where $GRU(\cdot)$ denotes standard GRU transition equations, $|\cdot|$ is the number of words, $w_{|c|}$ and $w_{|t_i|}$ is the last word of $c$ and $t_i$, respectively, $h_{|c|-1}$ and $h_{|t_i|-1}$ denote the hidden unit for the corresponding previous word, and $\theta_{c}$ and $\theta_{X}$ contain all the parameters in the claim and post encoder, respectively. 


\subsection{MIL-based Bottom-up Model} 
Subtree structure embeds relative stance patterns which are closely correlated to the claim (i.e., root node) veracity, e.g., ``$supp \to deny$" may be more frequently observed in a false rumor claim than that in a true claim, and false claims normally spark more denying stances in the posts than true claims do.  
The core idea of MIL-based bottom-up model is to infer the post stance supervised by the claim veracity annotations, while the prediction of both stance and veracity is jointly learned based on a recursive neural with bottom-up tree structure. 
For example, for the False Rumor bag in Figure~\ref{fig:fig1}, we could infer the posts, e.g., ``No he has not. The man himself confirmed not true" and ``No he has not", as holding the denying stance due to their salient opinion refuting the false rumor claim, and then deduce the stances of the connected posts with the help of the featured edges till the stances of all posts are determined following the propagation structure. 
The overall structure of our proposed bottom-up model is illustrated in Figure~\ref{fig:Model_bu}.

\textbf{Binary Stance Classification}. 
The RvNN model~\cite{ma2018rumor} assumes that the subtrees with similar contexts, e.g., subtrees with a supportive parent and denial children alike, can be represented with a similar vector of features that stand for the stances of the leading node 
and obtained by aggregating the post representations along different branches in the subtree. 
So, in a binary stance classifier $k$\footnote{$k$ is ranged in $[0, K-1]$.}, we obtain the stance representation for each post by synchronously aggregating the information from all its children nodes, following the similar bottom-up tree representation learning algorithm proposed in~\cite{ma2018rumor}. 
Specifically, we map the hidden vector $h_j$ of each node $j$ into a context vector $\tilde{h}_j$ by recursively combining the information of its children nodes $\mathcal{C}(j)$ and itself in each subtree: 
\begin{equation}\label{equ:BottomupPostPropagation}
    \tilde{h}_{j}^{k} = RvNN(h_{j}^{k}, h_{\mathcal{C}(j)}^{k}, \theta^{k})
\end{equation}
where $RvNN(\cdot)$ denote the bottom-up RvNN transition function~\cite{ma2018rumor}, $\mathcal{C}(j)$ represents the children nodes set of $j$ 
and $\theta^{k}$ represent all the parameters of RvNN.

We then use a fully-connected softmax layer to predict the stance probability of $t_j$ towards the claim vector $h_c^k$ related to classifier $k$: 
\begin{equation}\label{equ:postdistributions}
p_{j}^{k} = softmax(W^k_o\tilde{h}_{j}^{k} + W^k_c{h}_{c}^{k} + b^k_o)
\end{equation}
where $W^k_o$, $W^k_c$ and $b^k_o$ are the weights and bias, the probability values $\{p_{1}^{k},p_{2}^{k},\cdots,p_{T}^{k}\}$ can be considered as the stance-indicative probabilities of all the nodes in the tree inferred from the claim veracity. This is intuitive because the mapping between veracity and stance has been significantly reduced to more straightforward correspondences via the binarization of the classes, and the model can focus more on a salient subset of veracity-stance target class pairs such as F-D, T-S, U-Q and N-C, because of the close veracity-stance correlations in these pairs~\cite{mendoza2010twitter}. Meanwhile, the model would not simply disregard learning from other pairs by using an attentive process to automatically adjust the weights of the binary models corresponding to the pairs (see Section~\ref{sect:aggreg} for the details of binary models aggregation).  

\textbf{Binary Rumor Classification.} Based on the RvNN model, we aggregate stance indicative representations of the posts along the bottom-up structure into the global representation of a tree. Specifically, on top of the obtained stance distributions of the tree nodes from the binary stance classifier, we define a function to aggregate the stances to predict the veracity of the claim, following the intuition that the systematic synergy of user stances in the propagation can eventually support or refute the claim as to its veracity.  
To this end, we propose a 
\textit{bottom-up Tree Attention Mechanism} to selectively attend over specific stances expressed in more important posts from bottom to up recursively. In each loop of the recursion, 
let $\mathcal{S}(i)$ denotes the set of subtree nodes with root at $i$, 
the stance of node $i$ is updated as the aggregated stance in a subtree: 
\begin{equation}\label{equ:hierarchicalattention}
\begin{split}
    & \alpha_{j}^{k} = \frac{exp(\tilde{h}_{j}^{k} \cdot {h_{c}^{k}}^\top)}{\sum_{j \in \mathcal{S}(i)}exp(\tilde{h}_{j}^{k} \cdot {h_{c}^{k}}^\top)} \\
    & \tilde{p}_{i}^{k} = \sum_{j \in \mathcal{S}(i)}\alpha_{j}^{k} \cdot \tilde{p}_{j}^{k} 
\end{split}
\end{equation}
where $\alpha_{j}^{k}$ denotes the attention coefficient for each node $j \in \mathcal{S}(i)$, $\tilde{h}_{j}^{k}$ and $h_{c}^{k}$ respectively denote the hidden vector of post $j$ and claim $c$, and $\tilde{p}_{j}^{k}$ is the aggregated subtree stance. It is worth noting that Eq.~\ref{equ:hierarchicalattention} is inherently recursive following the bottom-up process in a sense that $\tilde{p}_{j}^{k} = p_{j}^{k}$ when $j$ is a leaf node, and when $i$ is the root node (i.e., $t_i=c$), we denote $\tilde{p}_{i}^{k} = \tilde{y}_c^k$. 

\subsection{MIL-based Top-Down Model}\label{section:TopDown}
The structure of top-down tree can capture complex stance patterns that model how information flows from source post to the current node. For example, a $``supp \to comm \to supp"$ path may be more common in true rumors than that in false rumors, and true rumor claims tend to spark more supportive posts than false claims. 
Similarly, the core idea of MIL-based top-down model is to infer the post-level stance along propagation path based on claim-level veracity label, while the prediction of both stance and veracity takes top-down structure. 
The overall structure of top-down model is shown in Figure~\ref{fig:Model_td}.

\textbf{Binary Stance Classification.}  Following information diffusion, we assume that the non-leaf stance features can be diffused synchronously to all its children nodes until the stance reaches the leaf nodes. So, in each binary classifier $k$, we can obtain the stance representation for each post by aggregating the post information along the propagation path starting from the source to current node.  
Based on the top-down representation learning algorithm proposed in~\cite{ma2018rumor}, 
we map each node $h_j$ into a context vector $\tilde{h}_j$ by recursively combining information of its parent node $P(j)$ and itself in each step\footnote{The RvNN-based representation learning in Eq.~\ref{equ:BottomupPostPropagation} and Eq.~\ref{equ:GRUPostPropagation} can be replaced with other state-of-the-art graph neural representation algorithms such as GCN~\cite{bian2020rumor}, tree-LSTMs~\cite{zhu2015long,tai2015improved} and PLAN~\cite{khoo2020interpretable}}:
\begin{equation}\label{equ:GRUPostPropagation}
    \tilde{h}_{j}^{k} = RvNN'(h_{j}^{k}, h_{P(j)}^{k}, \theta^k)
\end{equation}
where $RvNN'(\cdot)$ denotes the top-down-based RvNN transition function~\cite{ma2018rumor}, and $\theta^k$ represent all the corresponding parameters. 

We then use a fully-connected softmax layer to predict the stance probability of $t_j$ towards the claim vector $h_c^k$ related to classifier $k$: \begin{equation}\label{equ:postdistributions}
p_{j}^{k} = softmax(W^k_o\tilde{h}_{j}^{k} + W^k_c{h}_{c}^{k} + b^k_o)
\end{equation}
where $W^k_o, W^k_c{h}$ and $b^k_o$ are the weights and bias of prediction layer. Then, the stance probabilities of individual posts $\{p_{1}^{k},p_{2}^{k},\cdots,p_{T}^{k}\}$ 
in the top-down tree can be computed following the similar intuition as Eq.~\ref{equ:postdistributions}. 

\textbf{Binary Rumor Classification.} To aggregate top-down stance tree into claim veracity, we propose to attend over evidential stances along each propagation path as well as selecting more evidential paths for rumor veracity prediction. 
For this purpose, we design a \textit{top-down Tree Attention Mechanism} to aggregate the stances. 
Firstly, our model selectively attends on the evidential stance nodes in a path expressing specific attitude towards a claim. 
Let $r_l$ denote a propagation path from $c$ to $t_l$ (i.e., a leaf node), $\mathcal{P}(l)$ denotes node set along $r_l$, 
the aggregated stance for $r_l$ is obtained as follows:
\begin{equation}\label{equ:TopdownPathAttenion}
\begin{split}
    & \alpha_{j}^{k} = \frac{exp(\tilde{h}_{j}^{k} \cdot {h_c^{k}}^\top)}{\sum_{i \in \mathcal{P}(l)}exp(\tilde{h}_{i}^{k} \cdot {h_c^{k}}^\top)} \\
    & s_{l}^{k} = \sum_{j \in \mathcal{P}(l)}\alpha_{j}^{k} \cdot p_{j}^{k}
\end{split}
\end{equation}
where $\alpha_{j}^{k}$ denotes the attention coefficient of each node along $r_l$ and $s_{l}^{k}$ denotes the aggregated leaf node's stance probability along propagation path $r_l$. 

Secondly, for each path $r_l$, the information is eventually embedded into the hidden vector of the leaf nodes $\tilde{h}_{l}^{k}$. To further aggregate the path stance, we again adopt the tree attention mechanism to select more informative paths based on the leaf nodes. 
Let $\mathcal{K}(c)$ represent the leaf node set of the top-down tree rooted with claim $c$. The aggregated path stance representing the claim veracity can be computed as: 
\begin{equation}\label{equ:TopdownRumorPredict}
\begin{split}
    & \alpha_{l}^{k} = \frac{exp(\tilde{h}_{l}^{k} \cdot {h_c^{k}}^\top)}{\sum_{l \in \mathcal{K}(c)}exp(\tilde{h}_{l}^{k} \cdot {h_c^{k}}^\top)} \\
    & \tilde{y}_{c}^{k} = \sum_{l \in \mathcal{K}(c)}\alpha_{l}^{k} \cdot s_{l}^{k}
\end{split}
\end{equation}
where $\alpha_{l}^{k}$ denotes the attention coefficient of each leaf node and $\tilde{y}_{c}^{k}$ is the probability of rumor veracity which is aggregated from the stances of different paths. 

Although the discriminative tree attention for both MIL-based models aims to predict the claim veracity by recursively aggregating all the post stance, we can conjecture that the top-down model would be better. The hypothesis is that in the bottom-up case the stance is aggregated from local subtrees, and the context information is not fully considered compared with the top-down case where node stance is firstly aggregated through path locally then aggregate all paths globally.

\subsection{Binary Models Aggregation}\label{sect:aggreg}
It is intuitive that each binary classifier contributes differently to the final prediction, according to the different strength of the veracity-stance correlation it can capture. So, we design an attention mechanism to attend on the most reliable binary classifiers:
\begin{equation}\label{equ:betaattention}
    \begin{split}
    & h_a = GRU(w_{|c|}, h_{|c|-1}, \theta_a) \\
    & \beta_{k} = \frac{exp({h_a} \cdot {h_{c}^{k}}^\top)}{\sum_{k}exp({h_a} \cdot {h_{c}^{k}}^\top)}
    \end{split}
\end{equation}
where 
$\theta_a$ represents all the parameters inside the GRU encoder\footnote{Note that the GRU have a different set of parameters from Eq.~\ref{equ:GRUPostRepresentation}} and $h_c^k$ is the claim representation directly obtained from the $k$-th classifier.

\textbf{Stance Detection.} We regroup all the binary stance classifiers with the same stance type target $l_s \in \{S, D, Q, C\}$ into one set and then compute the final stance probability by:
\begin{equation}\label{equ:finalpostpredict}
  \hat{p}_{i,l_s} = \sum_{k \in U(l_s)}\beta_{k} \cdot p_{i}^{k} 
\end{equation}
where $U(l_s)$ represents the indicator set of the binary classifiers with $l_s$ as the target stance class, 
$p_{i}^{k}$ is the predicted stance probability of the post $t_i$ given by classifier $k$, and therefore $\hat{p}_{i,l_s}$ indicates the probability that the post $t_i$ is classified as stance $l_s$ weighted by the importance of the corresponding classifiers with $l_s$ as the target class. Note that these binary classifiers of the same target stance class are different models because each classifier is supervised by a different target veracity label of claim. Thus, the final probability distribution over all the stances can be obtained as $\hat{p}_i=[\hat{p}_{i,S}, \hat{p}_{i,D}, \hat{p}_{i,Q}, \hat{p}_{i,C}]$. 

\textbf{Rumor Verification.} We regroup all the binary claim veracity classifiers and put the classifiers with the same rumor class label $l_c\in\{N,T,F,U\}$ into one set. And then the claim veracity probability can be computed similarly as the weighted sum of all the binary classifiers’ outputs: 
\begin{equation}\label{equ:finalclaimpredict}
  \hat{y}_{c,l_c} = \sum_{k \in U(l_c)}\beta_{k} \cdot \tilde{y}_c^k 
\end{equation}
where $U(l_c)$ is the indicator set of the binary classifiers with $l_c$ as the target veracity class  
and $\tilde{y}_c^k$ is the predicted binary veracity class probability of the claim $c$ given by the classifier $k$. Note that the binary veracity classifiers with the same target veracity class are distinguishable by the different target stance classes they are associated with,  which give different stance probability predictions.
Thus, the final probability distribution over the veracity classes can be represented as $\hat{y}_c=[\hat{y}_{c,N}, \hat{y}_{c,T}, \hat{y}_{c,F}, \hat{y}_{c,U}]$. 

\subsection{Model Training}
To train each binary classifier, we transform the finer-grained veracity and stance labels into binary labels for ground-truth representation. For example, for the classifier with [T, S] veracity-stance pair as the target, the ground-truth veracity label of claim $y$ is represented as either `T' or `others', and the model outputs the stance for each post represented by a probability belonging to target class `S'. A variant of this similar setting is applicable to all the binary classifiers with different target classes. This yields the way for obtaining our binary ground truth:
\begin{equation}\label{equ:WeakClassifersLabelBags}
  y^{k}=
\begin{cases}
1& \text{if the target of classifier $k$ is the same as $y$}\\
0& \text{others}
\end{cases}  
\end{equation}
where $y \in [N,T,F,U]$ defined for the rumor verification task, and the ground truth refers to the veracity label of claims instead of stance due to the unavailability of label at post level.

\textbf{Binary MIL-based Classifiers Training.} We use the negative log likelihood 
as the loss function:
\begin{equation}\label{equ:lossfunction}
L_{bin} = -\sum_{k=1}^{K}\sum_{m=1}^{M} y_m^k*\log \hat{y}_m^k + (1-y_m^k)*\log(1-\hat{y}_m^k)
\end{equation}
where $y_m^k \in \mathbb [0, 1]$ indicating the ground truth of the $m$-th claim obtained in Eq.~\ref{equ:WeakClassifersLabelBags}, $\hat{y}_m^k$ is the predicted probability for the $m$-th claim in classifier $k$, $M$ is the total number of claims, and $K$ is the number of binary classifiers. 

\textbf{Aggregation Model Training.} We also utilize negative log likelihood loss function to train the aggregation model:
\begin{equation}\label{equ:finallossfunction}
L_{agg} = -\sum_{n=1}^{N} \sum_{m=1}^{M} y_{m,n}*\log \hat{y}_{m,n} + (1-y_{m,n})*\log(1-\hat{y}_{m,n})
\end{equation}
where $y_{m,n} \in [0, 1]$ is the binary value indicating if the ground-truth veracity class of the $n$-th claim is $m$, $\hat{y}_{m,n}$ is the predicted probability the  $n$-th claim belonging to class $m$, and $M$ is the number of veracity classes. 

All the parameters are updated by back-propagation~\cite{collobert2011natural} with Adam~\cite{DBLP:journals/corr/KingmaB14} optimizer. We use pre-trained GloVe Wikipedia 6B word embeddings~\cite{pennington2014glove} present on input words, set $d$ to 100 for word vectors
and empirically initialize the learning rate as 0.001. The training process ends when the loss value converges or the maximum epoch number is met\footnote{We set the number of maximum epoch as 150 in our experiment.}. The weighted aggregation model is trained after all the weak classifiers are well trained. 

\section{Experiments and Results}
\begin{table}[t]
  \centering
  \small
  \caption{Statistics of rumor datasets for model training.}
    \begin{tabular}{l|ccc}
    \toprule
    Statistics & Twitter15 & Twitter16  & PHEME \\
    \midrule
    \midrule
    \# of claim & 1,308  & 818   & 6,425  \\
    \# of Non-rumor & 374 (28.6\%) & 205 (25.1\%) & 4,023 (62.6\%)  \\
    \# of False-rumor & 370 (28.3\%) & 207 (25.3\%) & 638 (9.9\%)  \\
    \# of True-rumor & 190 (14.5\%) & 205 (25.1\%) & 1,067 (16.6\%) \\
    \# of Unverified-rumor & 374 (28.6\%) & 201 (24.5\%) & 697 (10.8\%) \\
    \midrule
    \# tree nodes & 68,026 & 40,867 & 383,569 \\
    \# of Avg. posts/tree & 52 & 50  & 6 \\
    \# of Max. posts/tree & 814 & 757 & 228 \\
    \# of Min. posts/tree & 1 & 1 & 3 \\
    \bottomrule
    \end{tabular}%
  \label{tab:TrainData}%
\end{table}%

\begin{table}[t]
  \centering
  \small
  \caption{Statistics of the datasets for model testing.}
    \begin{tabular}{l|cc}
    \toprule
    Statistics & RumorEval2019-S & SemEval8 \\
    \midrule
    \midrule
    \# of claim & 425 & 297 \\
    \# of Non-rumor & 100 (23.53\%) & —— \\
    \# of False-rumor & 74 (17.41\%) & 62 (20.8\%) \\
    \# of True-rumor & 145 (34.12\%) & 137 (46.1\%) \\
    \# of Unverified-rumor & 106 (24.94\%) & 98 (33.0\%)\\
    \midrule 			
    \# posts of Support & 1320 (19.65\%) & 645 (15.1\%)\\
    \# posts of Deny & 522 (7.77\%) & 334 (7.8\%)\\
    \# posts of Question & 531 (7.90\%) & 361 (8.5\%) \\
    \# posts of Comment & 4,345 (64.68\%) & 2,923 (68.6\%) \\
    \midrule
    \# tree nodes & 6,718 & 4,263 \\
    \# Avg. posts/tree & 16 & 14\\
    \# Max. posts/tree & 249 & 228\\
    \# Min. posts/tree & 2 & 3\\ 
    \bottomrule
    \end{tabular}%
  \label{tab:TestData}%
\end{table}%

\subsection{Datasets and Setup}
For experimental evaluation, we refer to rumor and stance dataset with propagation structure.  
For model training, since only rumor labels at claim-level are required, we refer to three public tree-based benchmark datasets for rumor verification on Twitter, 
namely Twitter15, Twitter16~\cite{ma2017detect} and PHEME\footnote{This PHEME dataset is not the one used for stance detection~\cite{zubiaga2016analysing}~\cite{derczynski2015pheme}. It is defined specifically for rumor detection task: \url{https://figshare.com/articles/PHEME_dataset_of_rumours_and_non-rumours/4010619}.}. In each dataset, every claim is annotated with one of the four veracity classes (i.e., Non-rumor, True, False and Unverified) and the post-level stance label is not available. We filter out the retweets since they simply repost the claim text. 

For model testing, since both post-level stance and claim level veracity are required, we resort to two rumor stance datasets collected from Twitter with stance annotations, 
namely RumorEval2019-S~\cite{gorrell2019semeval}\footnote{We only use the posts from Twitter in PhemeEval2019 and discard Reddit data. Note that although the original RumorEval2019-S and SemEval8 datasets contain some same claims, we further extend RumorEval2019-S dataset by including a more challenging but general case to evaluate the generalization power of our methods.} and SemEval8~\citep{derczynski2017semeval, zubiaga2016pheme}. The original datasets were used for joint detection of rumors and stances where each claim is annotated with one of the three veracity classes (i.e., true-rumor, false-rumor and unverified-rumor), and each responsive post is annotated as the attitude expressed towards the claim (i.e., agreed, disagreed, appeal-for-more-information, comment). 
We further  
augmented RumorEval2019-S dataset by collecting additional 100 non-rumor claims together with their relevant posts following the method described by~\citet{zubiaga2016analysing}. We asked three annotators independent of this work to annotate the post stance. 
For stance class, we convert the original labels into the [S, D, Q, C] set based on a set of rules proposed in~\cite{lukasik2016hawkes}. Table~\ref{tab:TrainData}--\ref{tab:TestData} display the data statistics. 

More specifically, 
when testing on RumorEval2019-S dataset, we train 16 binary classifiers in total considering that there are 4 veracity and 4 stance categories to be determined. And we train 12 binary stance classifiers for testing on SemEval-8 dataset since it contains 3 veracity and 4 stance categories. 
We hold out 20\% of the test datasets as validation datasets for tuning the hyper-parameters. 
Due to the imbalanced rumor and stance class distribution, accuracy is not sufficient for evaluation~\cite{zubiaga2016stance}. We use AUC, micro-averaged and macro-averaged F1 score, and class-specific F-measure as evaluation metrics. We implement all the neural models with Pytorch. 

\begin{table*}[t]
  \centering
  \caption{Results of stance detection.
  }  
  
    \begin{tabular}{l|lll|llll||lll|llll}
    \toprule
    Dataset & \multicolumn{7}{|c||}{RumourEval2019-S} & \multicolumn{7}{|c}{SemEval8} \\ \midrule
    \multirow{2}{*}{Method} & & & &  S & D & Q & C 
                   & & & & S & D & Q & C \\  \cline{5-8} \cline{12-15}
				& AUC & MicF & MacF & $F_1$ & $F_1$ & $F_1$ & $F_1$ 
				& AUC & MicF & MacF & $F_1$ & $F_1$ & $F_1$ & $F_1$ \\
     \midrule
     Zero-Shot & --- & 0.369 & 0.324  & 0.301  & 0.168  & 0.342  & 0.486 & --- & 0.383 & 0.344 & 0.278 & 0.162 & 0.480 & 0.456 \\
     Pre-Rule & --- & 0.605  & 0.478  & 0.657   & 0.419  &  ---  & --- & --- & 0.429 & 0.389 & 0.432 & 0.644 &  --- & --- \\
     C-GCN & 0.633 & 0.629 & 0.416 & 0.331 & 0.173 & 0.429 & 0.730 & 0.610 & 0.625 & 0.411 & 0.327 & 0.161 & 0.430 & 0.728\\
     \midrule
     BrLSTM(V) & 0.710 & 0.660 & 0.420 & 0.460 & 0.000 & 0.391 & 0.758 & 0.676 & 0.665 & 0.401 & 0.493 & 0.000 & 0.381 & 0.730 \\ 
     BiGRU(V) & 0.700 & 0.630 & 0.417 & 0.392 & 0.162 & 0.360 & 0.754 & 0.660 & 0.633 & 0.416 & 0.460 & 0.168 & 0.328 & 0.708\\
     MT-GRU(V) & 0.714 & 0.636 & 0.432 & 0.313 & 0.156 & 0.506 & 0.748 & 0.669 & 0.630 & 0.413 & \textbf{0.498} & 0.116 & 0.312 & 0.729\\
     \midrule
     \midrule
     TD-MIL(V) & 0.712 & 0.650 & 0.432 & 0.438 & 0.156 & 0.408 & 0.688 & 0.668 & 0.626 & 0.416 & 0.473 & 0.127 & \textbf{0.463} & 0.602\\
     BU-MIL(V) & 0.710 & 0.630 & 0.431 & \textbf{0.485} & 0.166 & 0.396 & 0.688 & 0.669 & 0.623 & 0.415 & 0.470 & 0.128 & 0.460 & 0.602\\
     \midrule
     \textbf{TD-MIL(T15)} & 0.706 & 0.668 & 0.427 & 0.339 & 0.173 & 0.444 & 0.752 & 0.663 & 0.642 & 0.418 & 0.330 & 0.174 & 0.420 & 0.750 \\
     \textbf{TD-MIL(T16)} & 0.713 & 0.665 & \textbf{0.436} & 0.350 & \textbf{0.182} & 0.446 & 0.758  & 0.660 & \textbf{0.671} & 0.421 & 0.334 & 0.173 & 0.422 & 0.754 \\
     \textbf{TD-MIL(PHE)} & \textbf{0.722} & \textbf{0.691} & 0.434 & 0.344 & 0.179 & \textbf{0.467} & \textbf{0.767} & \textbf{0.669} & 0.651 & \textbf{0.426} & 0.335 & \textbf{0.175} & 0.430 & \textbf{0.763} \\
     \midrule
     \textbf{BU-MIL(T15)} & 0.706 & 0.662 & 0.428 & 0.341 & 0.173 & 0.436 & 0.756 & 0.661 & 0.638 & 0.415 & 0.326 & 0.168 & 0.420 & 0.748 \\
     \textbf{BU-MIL(T16)} & 0.701 & 0.660 & 0.426 & 0.340 & 0.170 & 0.438 & 0.749 & 0.659 & 0.637 & 0.416 & 0.324 & 0.169 & 0.419 & 0.753 \\
     \textbf{BU-MIL(PHE)} & 0.707 & 0.665 & 0.432 & 0.344 & 0.174 & 0.445 & 0.762 & 0.666 & 0.642 & 0.420 & 0.329 & 0.169 & 0.423 & 0.758 \\
    \bottomrule
    \end{tabular}%
  \label{tab:StanceResult}%
\end{table*}%

\subsection{Stance Detection Performance}
Since our stance detection model is weakly supervised by coarse label (i.e., claim veracity) instead of explicit post-level stance label, we choose to compare with both unsupervised methods and supervised methods in the following: 
%
(1) \textbf{Zero-Shot}~\cite{allaway2020zero}: A pre-trained stance detection method that captures relationships between topics. 
(2) \textbf{Pre-Rule}~\cite{kobbe2020unsupervised}: An unsupervised method designed for detecting support and deny stance by referring to some pre-defined rules. 
(3) \textbf{C-GCN}~\cite{wei2019modeling}: An unsupervised graph convolutional network that classifies the stances by modeling tweets with conversation structure. 
(4) \textbf{BrLSTM}~\cite{kochkina2017turing}: An LSTM-based model that models the conversational branch structure of tweets to detect stance. 
(5) \textbf{BiGRU}~\cite{augenstein2016stance}: A bidirectional RNN-based tweet stance model which considered the bidirectional contexts between target and tweet. We replaced the original LSTM units with GRU for fair comparisons. 
(6) \textbf{MT-GRU}~\cite{ma2018detect}: 
A multi-task learning approach based on GRU for joint detection of rumors and stances by capturing the both shared and task-specific features. 
%
\textbf{TD/BU-MIL(\textsc{DateSet})} is our proposed MIL-based top-down/bottom-up model using the veracity labels in \textsc{DateSet} for the weak supervision\footnote{\textbf{T15}, \textbf{T16} and \textbf{Phe} are the short-forms of Twitter15, Twitter16 and PHEME datasets, respectively. 
In order to make fair comparison with the supervised stance detection models that are trained with validation sets, here \textbf{V} denotes the variants of our models trained with the validation sets although only veracity labels are used.} 


In Table~\ref{tab:StanceResult}, we use the open source of Zero-Shot and Pre-rule, which does not report AUC. Zero-shot, Pre-Rule and C-GCN are models without the need of annotated data for stance detection, while BrLSTM, BiGRU and MT-GRU 
are three popular supervised stance detection baseline models. 
To train the supervised models, we use the validation datasets left out from RumorEval2019-S and SemEval-8 in the same way as our fully supervised variants TD/BU-MIL(V) does for fair comparison. This is because there is no stance annotations in the training set. 

The first group refers to unsupervised baselines. Zero-Shot and Pre-rule perform worse than other methods, because they are pre-trained models based on the out-of-domain data from their original papers that cannot generalize well to Twitter data. The results on Q and C by Pre-Rule are absent since the pre-defined linguistic rules are designed for identifying the stance of Support and Deny only. Propagation-based method C-GCN performs better because it capture structural information in propagation by modeling the neighbors of each tweet. 

The second group considers supervised baselines. BrLSTM improves unsupervised baselines with a large margin in terms of Micro-F1, 
because BrLSTM is focused on modeling propagation structure while both BiGRU and MT-GRU are sequential models.
But BrLSTM is poor at classifying denial stance since it is data-driven and the proportion of the stance is small in the training data. 
Our method TD-MIL(V) achieves comparable Micro-F1 and Macro-F1 scores as BiGRU, 
indicating that our MIL-based method has the potential to surpass supervised models. This is because TD-MIL(V) considers the information propagation patterns in the whole propagation tree while BiGRU only makes limited comparisons between the target veracity class and the unstructured posts.

Our MIL-based method outperforms all the baselines when training data is large enough. For example, BU/TD-MIL(Phe) performs better than its counterparts trained on Twitter15/16 datasets. This is because PHEME contains more claims than Twitter15/16 datasets for weak supervision. We conjecture that the performance of our method can be further improved when trained on larger datasets. 
\begin{table*}[htbp]
  \centering
  \caption{Results of rumor verification.
  }
    \begin{tabular}{l|lll|llll||lll|lll}
    \toprule
    Dataset & \multicolumn{7}{|c||}{RumorEval2019-S} & \multicolumn{6}{|c}{SemEval8} \\ \hline
    \multirow{2}{*}{Method} & & &  &T & F & U & N 
                            & & & & T & F & U\\  \cline{5-8} \cline{12-14}
				& AUC & MicF & MacF & $F_1$ & $F_1$ & $F_1$ & $F_1$
				& AUC & MicF & MacF & $F_1$ & $F_1$ & $F_1$ \\
    \midrule
    GCAN & 0.693 & 0.645 & 0.253 & 0.249 & 0.310 & 0.113 & 0.339 & 0.688 & 0.645 & 0.255 & 0.241 & 0.326 & 0.198 \\
    PPC & 0.672 & 0.632 & 0.250 & 0.244 & 0.296 & 0.114 & 0.346 &  0.673 & 0.642 & 0.249 & 0.237 & 0.289 & 0.221 \\
    TD-RvNN & 0.880 & 0.743 & 0.699 & 0.713 & 0.631 & 0.660 & 0.792 & 0.882 & 0.728 & 0.689 & 0.702 & 0.619 & 0.745 \\
    BU-RvNN & 0.865	& 0.720 & 0.723 & 0.746 & 0.641 & 0.696 & 0.806 & 0.870 & 0.708 & 0.684 & 0.708 & 0.620 & 0.723 \\
    H-GCN & 0.690 & 0.534 & 0.418 & 0.712 & 0.180 & 0.371 & 0.409 & 0.675 & 0.530 & 0.413 & 0.355 & 0.16 & 0.724 \\
    \midrule
    MTL2 (V) & 0.683 & 0.653 & 0.430 & 0.622 & 0.279 & 0.352 & 0.457 & 0.680 & 0.651 & 0.433 & 0.640 & 0.289 & 0.372 \\
    MT-GRU (V) & 0.704 & 0.768 & 0.452 & 0.462 & 0.298 & 0.373 & 0.452 & 0.701 & 0.761 & 0.428 & 0.639 & 0.254 & 0.391\\
    \midrule
    \midrule
    TD-MIL (V) & 0.685 & 0.678 & 0.450 & 0.667 & 0.329 & 0.376 & 0.428 & 0.680 & 0.621 & 0.436 & 0.650 & 0.274 & 0.384\\
    BU-MIL (V) & 0.682 & 0.679 & 0.448 & 0.668 & 0.326 & 0.373 & 0.428 & 0.680 & 0.645 & 0.427 & 0.631 & 0.292 & 0.360\\
    \midrule
    \textbf{TD-MIL (T15)} & \textbf{0.919} & 0.793 & \textbf{0.790} & 0.822 & \textbf{0.762} & 0.716 & 0.818 & \textbf{0.913} & 0.771 & 0.730 & 0.679 & \textbf{0.689} & 0.823\\
    \textbf{TD-MIL (T16)} & 0.914 & 0.792 & 0.764 & 0.796 & 0.740 & 0.719 & 0.812 & 0.899 & 0.785 & 0.725 & 0.668 & 0.682 & 0.825\\
    \textbf{TD-MIL (Phe)} & 0.917 & \textbf{0.809} & 0.776 & \textbf{0.826} & 0.659 & 0.669 & \textbf{0.852} & 0.908 & \textbf{0.798} & \textbf{0.741} & \textbf{0.741} & 0.672 & 0.810\\
    \midrule
    \textbf{BU-MIL (T15)} & 0.899 & 0.769 & 0.780 & 0.794 & 0.688 & 0.770 & 0.819 & 0.887 & 0.752 & 0.724 & 0.670 & 0.680 & 0.822\\
    \textbf{BU-MIL (T16)} & 0.902 & 0.776 & 0.760 & 0.780 & 0.664 & \textbf{0.780} & 0.810 & 0.893 & 0.756 & 0.721 & 0.663 & 0.676 & \textbf{0.826}\\
    \textbf{BU-MIL (Phe)} & 0.904 & 0.776 & 0.763 & 0.793 & 0.666 & 0.770 & 0.833 & 0.902 & 0.763 & 0.729 & 0.728 & 0.649 & 0.809\\
    \bottomrule
    \end{tabular}%
  \label{tab:RumorResult}%
\end{table*}%


\subsection{Rumor Verification Performance}
We compare our methods with the following state-of-the-art rumor verification baselines: 
(1) \textbf{TD-RvNN} and (2) \textbf{BU-RvNN}~\cite{ma2018rumor}: A tree-structured recursive neural networks for rumor verification with top-down and bottom-up propagation structure. 
%
%
(3) \textbf{H-GCN}~\cite{wei2019modeling}: A hierarchical multi-task learning framework for jointly predicting rumor and stance with graph convolutional network.
(4) \textbf{GCAN}~\cite{lu2020gcan}: A graph-aware co-attention model utilizing retweet structure to verify the source tweet. 
(5) \textbf{PPC}~\cite{liu2018early}: a propagation-based early detection model utilizing user information and retweets. 
%
%
(6) \textbf{MT-GRU}~\cite{ma2018detect}: 
A multi-task learning approach to jointly detect rumors and stances by capturing both shared and task-specific features. 
(7) \textbf{MTL2}~\cite{kochkina2018all}: 
A sequential approach sharing a LSTM layer between the tasks, which is followed by a number of task-specific layers for multi-task outputs. 
\textbf{TD/BU-MIL(\textsc{DateSet})} is our MIL-based methods for rumor verification. 

In Table~\ref{tab:RumorResult}, we only report the best result of supervised rumor verification methods in the first group across different training datasets. MT-GRU and MTL2 are multi-task models which require training with both claim and post labels. So in our case, they are trained on the validation datasets because training datasets only have claim-level label while the validation dataset has both veracity and post stance labels. For fair comparisons with supervised baselines, TD/BU-MIL(V) are trained on the same validation datasets. To further improve and generalize our method, we train all the MIL-based models with larger datasets, i.e., T15, T16 and Phe.

Among the first group that relates to structured supervised baselines, we observe that GCAN, PPC and H-GCN perform worse than the other systems, because they only consider local structure such as directly connected neighborhood. PPC performs poorly because the number of user nodes in their model include both reply and retweet that can provide user features, which is much more than the number of reply nodes in our model.  
In contrast, TD-RvNN and BU-RvNN perform better because they model the global propagation contexts by aggregating the entire propagation information recursively. However, they perform worse than our MIL methods since our methods (i.e., TD-MIL(*) and BU-MIL(*)) not only use propagation information for tweets representation, but also aggregate stances with tree-based and MIL-based attention mechanism, which reduces the impact of noise stances. 

The second group consists of non-structured multi-task frameworks utilizing both post-level and claim-level labels. MT-GRU(V) and MTL2(V) get higher Micro-F1 and Macro-F1 score than our method TD-/BU- MIL(V). This is because they are both trained under the supervision of veracity and stance annotation whereas our method only utilizes veracity labels. 
Moreover, TD-MIL(V) shows comparable AUC and MacF scores with MTL2(V), 
suggesting the potential of our weakly supervised model compared to the supervised baselines. By increasing the training sets, our models precede the multi-task models as TD-/BU- MIL(T15/T16/Phe) do. 

Among the multi-task models, 
we observe that our MIL models are more effective than the non-structured baselines (e.g., MTL2, MT-GRU), because our MIL-based propagation models consider structural features. TD-MIL (*) outperforms BU-MIL (*) because TD-MIL (*) considers both local and global contexts during stance aggregation, 
which verifies our assumptions in Section~\ref{section:TopDown}. 


\begin{table}[t]
  \centering
  \caption{Ablation study results.}
    \begin{tabular}{l|rrr||rrr}
    \toprule
          & \multicolumn{3}{c||}{Rumor Result} & \multicolumn{3}{c}{Stance Result} \\
    \midrule
    Method & \multicolumn{1}{l}{AUC} & \multicolumn{1}{l}{MicF} & \multicolumn{1}{l||}{MacF} & \multicolumn{1}{l}{AUC} & \multicolumn{1}{l}{MicF} & \multicolumn{1}{l}{MacF} \\
    \midrule
    MIL-a & 0.892 & 0.759 & 0.736 & 0.672 & 0.643 & 0.430 \\
    \midrule
    TD-MIL-b & 0.912 & 0.802 & 0.746 & 0.701 & 0.658 & 0.426 \\
    TD-MIL-c & 0.903 & 0.805 & 0.738 & 0.696 & 0.653 & 0.420 \\
    \midrule
    BU-MIL-b & 0.901 & 0.752 & 0.743 & 0.698 & 0.647 & 0.419 \\
    BU-MIL-c & 0.903 & 0.749 & 0.742 & 0.687 & 0.645 & 0.419 \\
    \midrule
    \midrule
    TD-MIL & 0.917 & 0.809 & 0.776 & 0.722 & 0.691 & 0.434 \\
    BU-MIL & 0.904 & 0.776 & 0.763 & 0.707 & 0.665 & 0.432 \\
    \bottomrule
    \end{tabular}%
  \label{tab:ablationstudy}%
\end{table}%

\subsection{Ablation Study}
To evaluate the impact of each component, we perform ablation tests based on the best performed BU-/TD-MIL (Phe) models on RumorEval2019-S dataset by removing some component(s): 
1) \textbf{MIL-a}: replace all tree-based post encoder with non-structured post encoder and remove the tree-based recursive stance aggregation process;  
2) \textbf{TD/BU-b}: replace top-down/bottom-up post encoder with non-structured post encoder while keeping the tree-based attention mechanism;
3) \textbf{TD/BU-c}: replace top-down/bottom-up tree attention mechanism with the general dot product attention~\cite{vaswani2017attention} for stance aggregation.

As shown in Table~\ref{tab:ablationstudy}, MIL-a gets the lowest performance scores, where AUC/MicF/MacF decreases about 2.5\%/5.4\%/4\% for rumor verification and 5\%/4.8\%/0.4\% for stance detection, which demonstrates top-down/bottom-up tree structure is vital to our methods. Besides, the variants in terms of BU/TD-MIL-c result in the largest performance drops in both directions for the two tasks, indicating that discriminative tree-based attention mechanisms for stance aggregation play an important role in the our MIL method.

\subsection{Case Study}
\begin{figure}[t!]
\centering
\subfigure[True rumor case]{
\includegraphics[scale=0.312]{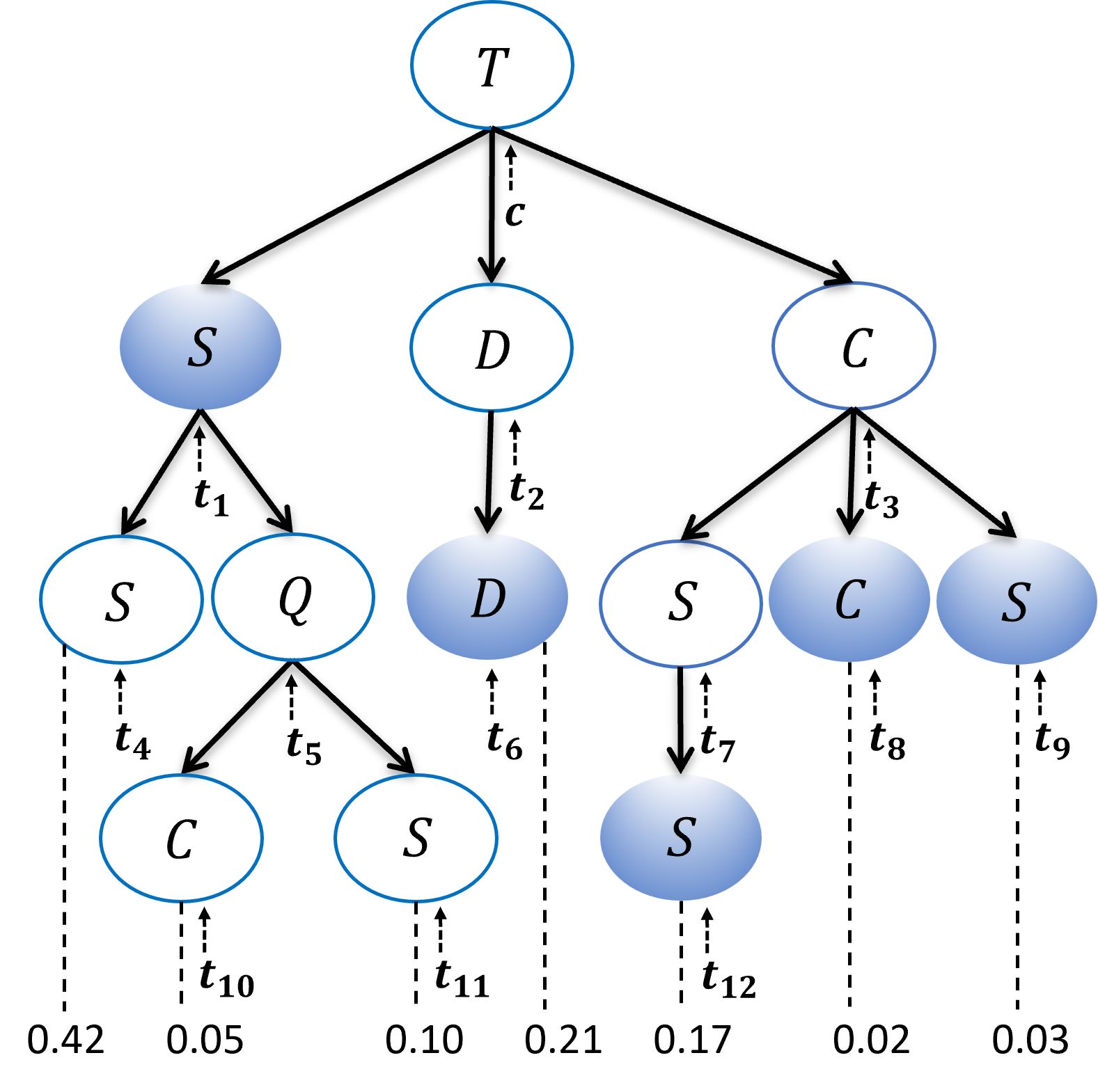}
\label{fig:TrueCase}}
\subfigure[False rumor case]{
\includegraphics[scale=0.312]{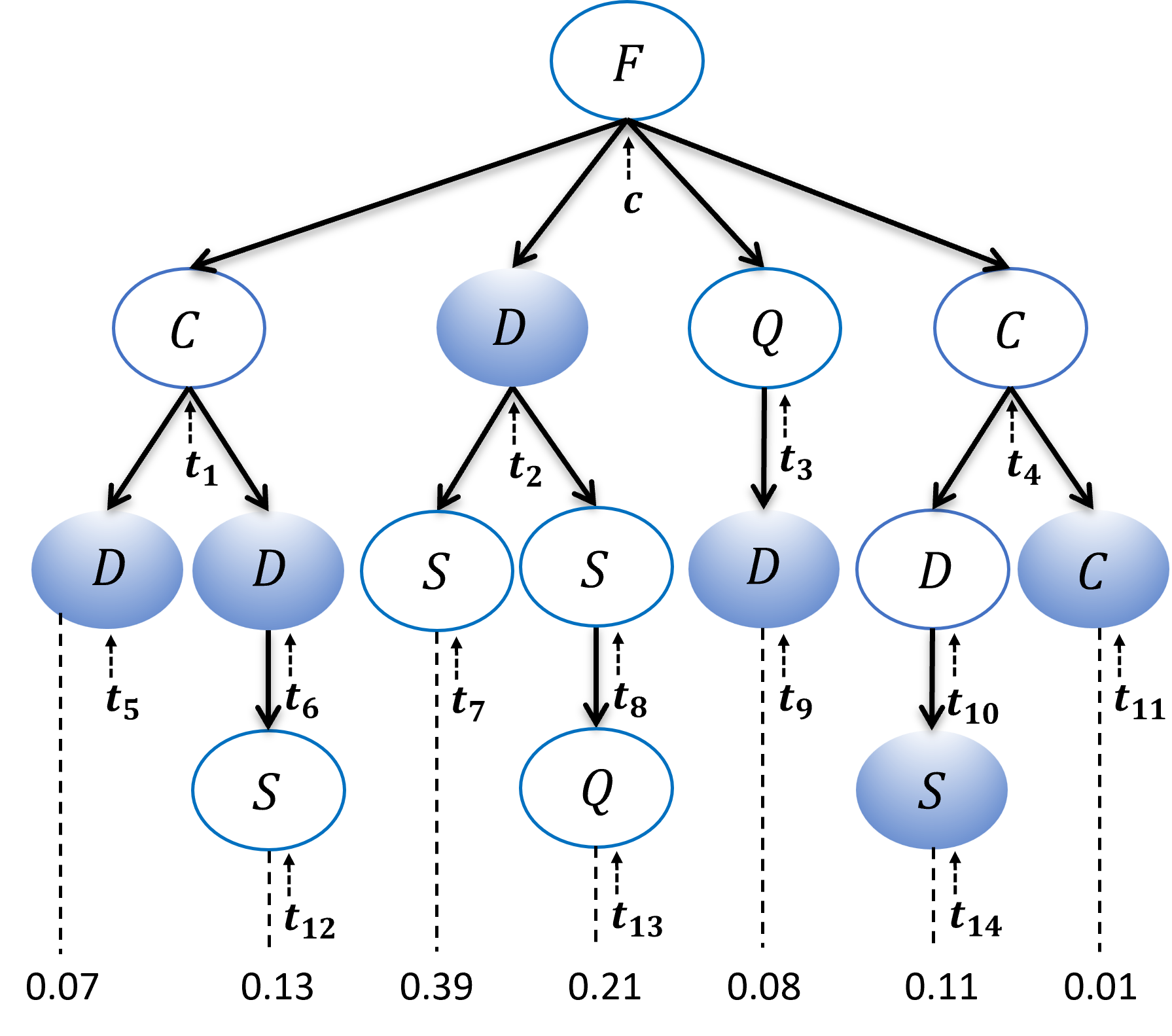}
\label{fig:FalseCase}}
\caption{Case study of tree-based attention mechanism.}\label{fig:CaseStudyTree}
\end{figure}

To get an intuitive understanding of the tree attention mechanism, we design an experiment to show the behavior of TD-MIL (Phe), due to its superior performance. Specifically, we sample two trees from RumorEval2019-S that the source claims have been correctly classified as True and False rumor, and display the posts' predicted stance results. We compute the average path/leaf nodes attention scores over all binary classifiers, mark the most important stance with solid blue oval for each propagation path and show the leaf nodes attention scores corresponding to the importance of each propagation path in Figure~\ref{fig:CaseStudyTree}. We observe that: 1) The ``support" posts mostly play an important role along each propagation path with True rumor as the target class. 2) The ``deny" posts contribute more in each propagation path with False rumor as the target class. 3) The model with true rumor as target attends more on $``support \to support"$ and $``deny \to deny"$ propagation patterns, which end with $t_4$ and $t_6$ respectively as shown in the figure. 4) The model with False rumor as target tends to capture the $``deny \to support"$ and $``comment \to deny \to support"$  propagation patterns which end with $t_7$ and $t_6$, respectively. These observations conform to our assumption and intuition that the tree propagation structure is vital in the joint task of rumor verification and stance detection. 

\begin{figure}[t!]
\centering
\subfigure[RumorEval2019-S]{
\includegraphics[scale=0.245]{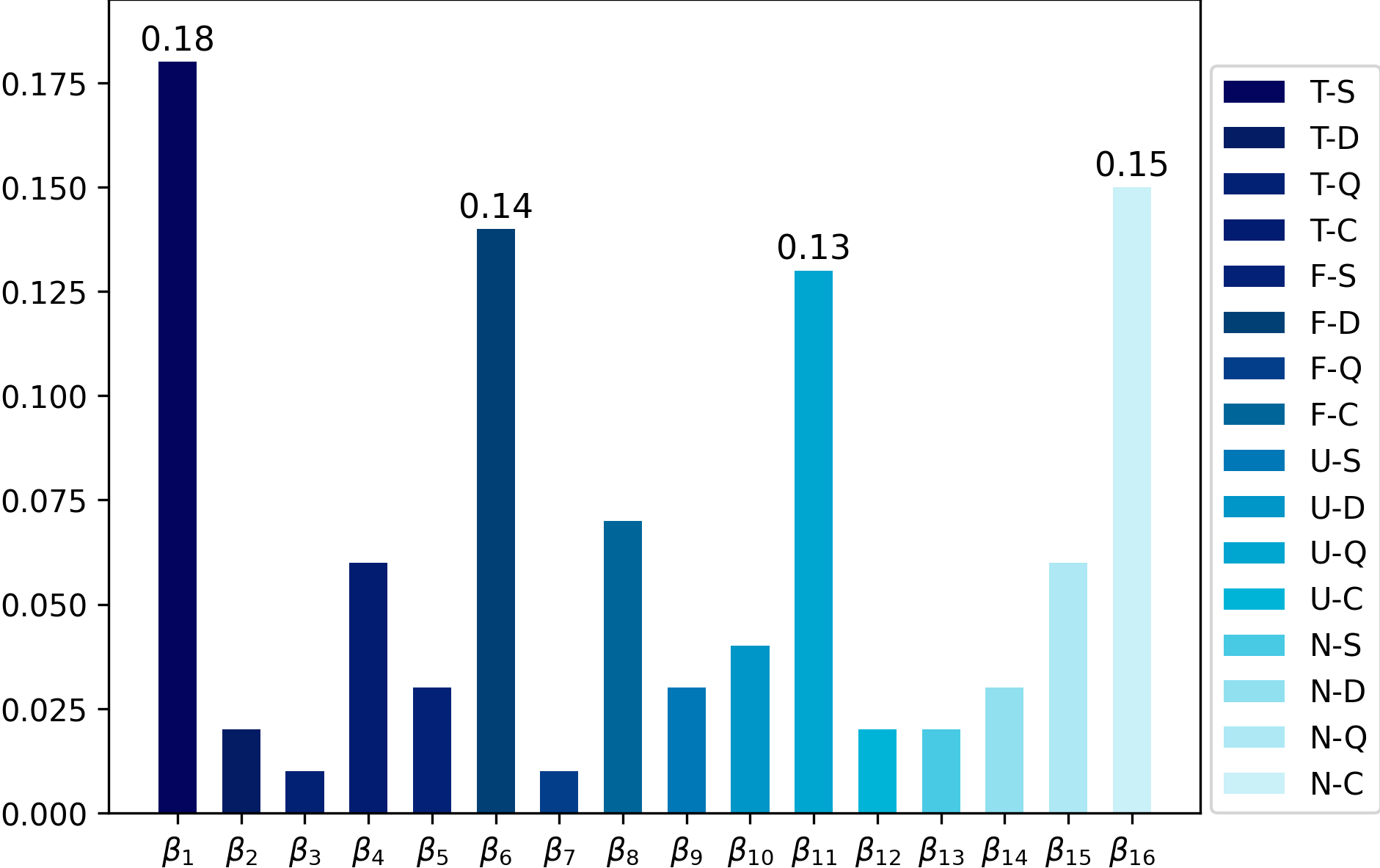}
\label{fig:RumorEval2019-SAttention}}
\subfigure[SemEval8]{
\includegraphics[scale=0.245]{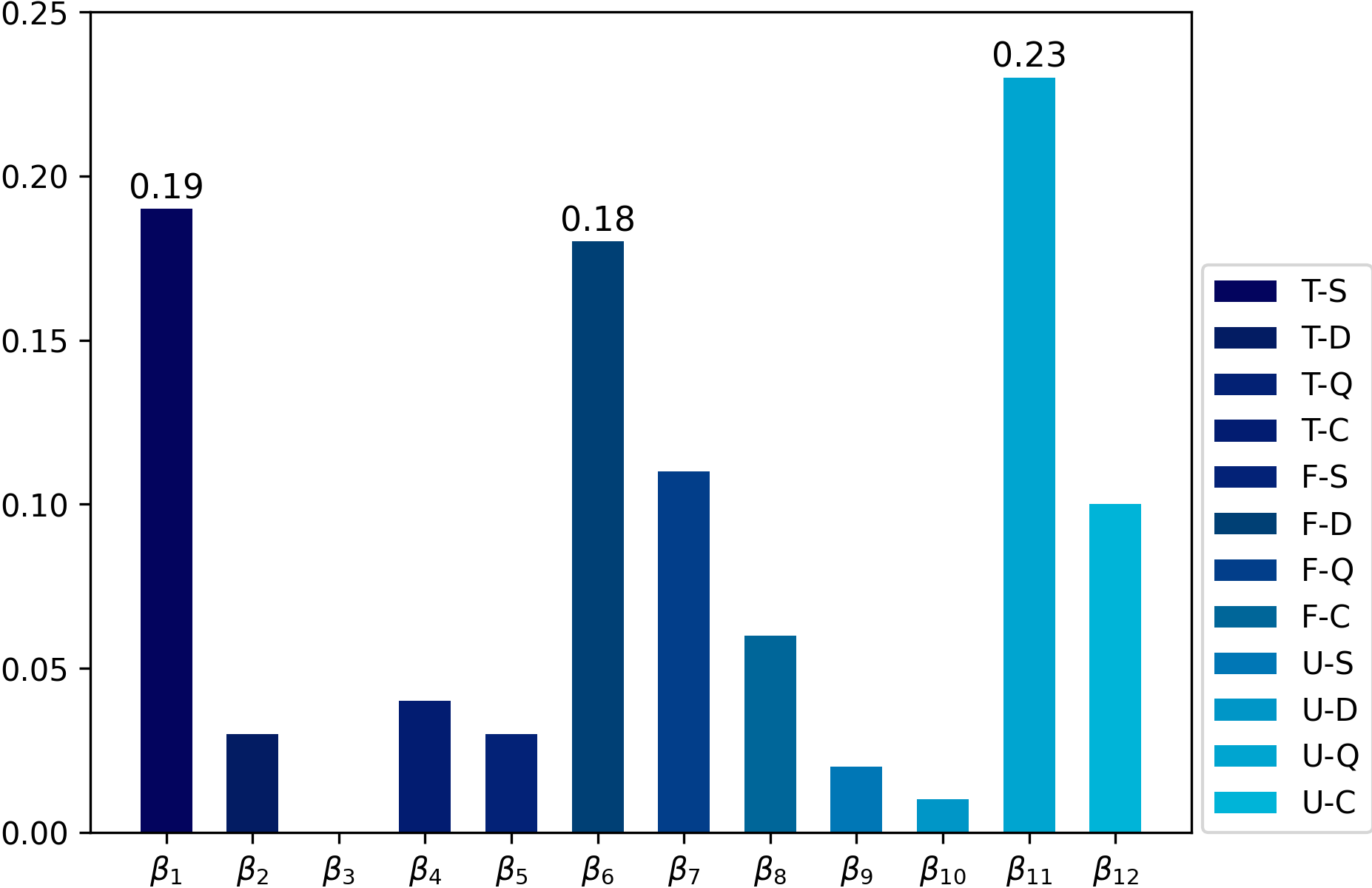}
\label{fig:SemEval8Attention}}
\caption{Average attention scores for binary classifiers obtained from Eq.~\ref{equ:betaattention}. 
}\label{fig:CaseStudyAttention}
\end{figure}

We also conduct experiments to show why the aggregation model can simultaneously enhance rumor verification and stance detection tasks. We randomly sample 100 claims from PHEME dataset, and then disclose the attention scores of all the binary classifiers obtained during the evaluations on RumorEval2019-S and SemEval8 datasets. The average attention scores over all the claims are shown in Figure~\ref{fig:CaseStudyAttention}. We observe that: 1) The top attention scores indicate a close correlation between the specific rumor veracity and stance category, which is compatible with previous findings~\cite{mendoza2010twitter}. For instance, the high values of $\beta_1$ and $\beta_6$ in Figure~\ref{fig:RumorEval2019-SAttention} 
suggest that true rumor and supportive stance in the posts are more closely correlated and likewise the false rumor and denial stance. 
2) The classifiers with lower attention suggest that the final prediction prediction is less affected by the corresponding veracity-stance relationship, indicating there is a weak correlation between rumor and stance for the current target class. 
For example, $\beta_2$ and $\beta_3$ in Figure~\ref{fig:SemEval8Attention} demonstrates that for the T-D and T-Q target pairs, veracity-stance correlation has weak influence on the prediction of the corresponding binary models. This seems to be consistent with the lower proportion of denying and questioning posts as shown in Table~\ref{tab:TestData}. 3) The rumor veracity can be generally better determined based on a combination of diverse stances instead of a single stance. For instance, support and comment stances combined can contribute more than others when true rumor being the target class, which can be reflected by the relatively high value when combining T-S ($\beta_1$) and T-C ($\beta_4$) model predictions as shown in Figure~\ref{fig:RumorEval2019-SAttention}.
4) Similarly, among the
classifiers with question stance as the target class, the T-Q 
model is generally less important than the other three (i.e., N-Q, F-Q and U-Q) which indicates the lower proportion of questioning posts in the true rumor.

\section{Conclusion}
We propose two tree-structured weakly supervised propagation models based on Multiple Instance Learning (MIL) framework for simultaneously verifying rumorous claims and detecting stances of their relevant posts. Our models are trained only with bag-level annotations (i.e., claim veracity labels), which can jointly infer rumor veracity and the unseen post-level stance labels. Our two novel tree-based stance aggregation mechanisms in the top-down and bottom-up settings achieve promising results for both rumor verification and stance detection tasks in comparison with state-of-the-art supervised and unsupervised models. 


\begin{acks}
This work was partially supported by HKBU One-off Tier 2 Start-up Grant (Ref. RCOFSGT2/20-21/SCI/004) and HKBU direct grant (Ref. AIS 21-22/02).

\end{acks}

\bibliographystyle{ACM-Reference-Format}
\bibliography{sample-base}

\end{document}